# Evaluating Financial Sentiment in the Age of AI

## Draft: July 17, 2026

Arslan Bisharat
Loyola University of Chicago, USA
marslan@luc.edu

Oudom Hean
College of Business and Challey Institute, North Dakota State University, USA
oudom.hean@ndsu.edu

**Abstract**

Financial sentiment measures are widely used in empirical finance, but it remains unclear whether general-purpose large language models (LLMs) improve on existing finance-specific methods. This paper evaluates twelve sentiment models, including dictionary-based methods, finance-specific transformers, and open-source LLMs, using two criteria: linguistic validity and economic validity. We find that general-purpose LLMs achieve classification performance comparable to finance-specific transformer models without task-specific fine-tuning. However, higher classification accuracy does not translate into stronger economic relationships. Several models produce sentiment measures that are significantly associated with earnings surprises, but none is significantly associated with next-day stock returns. Model performance is strongest for announcements with large earnings beats or misses and substantially weaker for announcements with more moderate earnings surprises. These findings suggest that financial sentiment captures information about firms' economic performance but has limited ability to explain short-run market reactions.

## 1. Introduction

Financial text has become an increasingly important source of information in finance. Researchers, traders, and investors routinely analyze corporate disclosures, earnings announcements, analyst reports, and financial news to examine investor sentiment, managerial tone, information asymmetry, and asset prices. Because sentiment is not directly observable, empirical studies rely on natural language processing (NLP) methods to transform financial text into quantitative measures that can be incorporated into statistical analyses. However, it remains unclear how financial sentiment models should be evaluated and whether stronger performance on linguistic benchmarks translates into greater usefulness for empirical finance.

Our paper evaluates financial sentiment models using two complementary criteria. The first is linguistic validity, which measures how closely model predictions agree with expert human annotations. We assess linguistic validity using the Financial PhraseBank dataset, one of the most widely used benchmark datasets in financial sentiment analysis. The second is economic validity, which measures whether model-generated sentiment is associated with economically meaningful outcomes. We evaluate economic validity using earnings announcements obtained from SEC 8-K filings and examine the relationships among sentiment, earnings surprises, and next-day stock returns.

Our analysis compares twelve sentiment models representing three generations of NLP methods: dictionary-based approaches, finance-specific transformer models, and modern open-source large language models. This comparison allows us to examine whether recent advances in language modeling improve the measurement of financial sentiment and, more importantly, whether those improvements extend beyond benchmark classification tasks.

The distinction between linguistic and economic validity is important because financial sentiment is a latent construct rather than a directly observable variable. A model may closely reproduce expert sentiment labels yet fail to capture information that is relevant to firms or financial markets. Conversely, a model that performs slightly worse on a benchmark dataset may generate sentiment measures that more accurately reflect firms' underlying economic conditions. Evaluating sentiment models solely on classification accuracy may therefore provide an incomplete assessment of their usefulness for empirical research.

Early studies of financial sentiment primarily relied on dictionary-based approaches, such as the Loughran and McDonald (2011) financial dictionary, to analyze corporate disclosures. More recently, advances in machine learning have led to the development of finance-specific language models, particularly FinBERT (Araci, 2019), which has become a widely used tool for financial sentiment analysis. These models generally outperform dictionary-based methods because they account for linguistic context and finance-specific terminology.

The rapid development of large language models raises a new question for empirical finance. Open-source and openly available general-purpose models, such as Llama, Qwen, Gemma, and Mistral, have demonstrated strong performance across a wide range of language tasks, often without task-specific training. These models are particularly attractive because they can be deployed locally, applied to large datasets at relatively low marginal cost, and used without transmitting potentially sensitive financial text to proprietary platforms. Their accessibility also enhances transparency and reproducibility by allowing analysts to document model versions, prompts, and inference procedures.

In this respect, open-source LLMs may provide a cost-effective, scalable, and reproducible alternative to both dictionary-based methods and finance-specific language models. Despite these potential advantages, limited evidence exists on whether open-source general-purpose LLMs produce more accurate or economically meaningful measures of financial sentiment than finance-specific models. It also remains unclear whether improvements in benchmark classification accuracy translate into stronger relationships with economically relevant outcomes.

We report three main findings. First, modern open-source LLMs achieve classification performance comparable to finance-specific transformer models despite requiring no task-specific fine-tuning. Second, although several models exhibit statistically significant associations with earnings surprises, none is significantly associated with next-day stock returns. Third, sentiment models are most effective at identifying announcements associated with large earnings beats and misses but provide considerably weaker signals for announcements with more modest earnings surprises.

We conduct an additional analysis by benchmarking these models against Claude, a leading closed-source frontier general-purpose LLM. Claude achieves performance comparable to that of the open-source LLMs, suggesting that current open-source models provide a cost-effective alternative for financial sentiment analysis without a meaningful loss in performance.

Our paper contributes to the literature in three ways. First, we introduce a framework that distinguishes between linguistic and economic validity when evaluating financial sentiment measures. Second, we provide one of the first comprehensive comparisons of modern open-source LLMs and finance-specific language models using both benchmark data and real-world earnings announcements. Finally, we demonstrate that improvements in benchmark classification accuracy do not necessarily translate into greater economic relevance, highlighting the importance of validating sentiment measures against economically meaningful outcomes.

The remainder of the paper is organized as follows. Section 2 reviews the related literature. Section 3 describes the data and sentiment models. Section 4 presents the empirical methodology. Section 5 reports the main results. Section 6 presents additional analysis by benchmarking the models against Claude, a leading closed-source frontier LLM. Section 7 discusses the implications for empirical finance research, and Section 8 concludes.

## 2. Literature Review

Financial sentiment has long been recognized as an important determinant of financial markets. Early studies show that information contained in news articles, corporate disclosures, and other textual sources affects investor expectations and asset prices (Baker and Wurgler, 2006; Tetlock, 2007; Tetlock et al., 2008). Since then, textual analysis has become a standard tool in empirical finance, with applications ranging from asset pricing and corporate disclosure to earnings announcements, analyst reports, and risk measurement. Because financial sentiment cannot be directly observed, researchers increasingly rely on computational methods to convert text into quantitative measures suitable for empirical analysis.

The earliest approaches to financial sentiment analysis relied on dictionary- and rule-based methods. General-purpose tools such as TextBlob and VADER provide accessible and computationally inexpensive ways to convert text into numerical sentiment scores (Hutto & Gilbert, 2014; Loria, 2018). However, these general-purpose methods often perform poorly in financial settings because many words have meanings that differ from their everyday usage. To address this limitation, Loughran and McDonald (2011) developed a finance-specific dictionary that substantially improved sentiment measurement in corporate filings. Although dictionary- and rule-based methods remain popular because they are transparent and computationally inexpensive, they have limited ability to capture word order and broader linguistic context, reducing their effectiveness in interpreting complex financial language.

Recent advances in machine learning have led to the development of transformer-based language models for financial text analysis. Building on the BERT architecture, finance-specific models such as FinBERT and related variants have become widely used in both academic research and industry applications (Araci, 2019; Devlin et al., 2019; Hazourli, 2022; Huang et al., 2023; Liu, et al. 2019; Pollé, 2021; Romero, 2021; Sanh et al., 2019; Yang et al., 2020). These models generally outperform dictionary-based methods on benchmark datasets because they capture contextual relationships among words and learn domain-specific linguistic patterns.

The rapid development of large language models has created new opportunities for financial text analysis. Large-scale language models can perform a variety of tasks through zero-shot or few-shot prompting without task-specific parameter updating (Brown et al., 2020). General-purpose models such as Llama, Qwen, Gemma, and Mistral demonstrate strong performance across a broad range of language tasks without requiring task-specific training. Their flexibility raises the possibility that finance-specific models may no longer be necessary for many applications.

Other recent studies have explored the application of machine learning and LLMs in accounting and finance. Kim et al. (2024) examine whether GPT can perform financial statement analysis comparable to that of professional analysts, while Siano (2025) demonstrates that LLMs substantially improve the extraction of information from earnings announcement disclosures. Jang and Wu (2026) demonstrate that LLMs can effectively analyze sentiment in non-English Management Discussion and Analysis (MD&A) disclosures, whereas Taraj and Wahlstrøm (2026) use machine learning to examine how narrative characteristics of earnings conference calls relate to market outcomes. Although recent studies have examined the use of proprietary LLMs, particularly GPT, for financial text analysis, evidence on the performance of open-source LLMs remains limited (see Hean et al., 2025; Kang et al., 2025).

## 3. Data

We use the Financial PhraseBank (Malo et al., 2014) to evaluate the ability of sentiment models to reproduce expert human judgments. The dataset consists of financial news sentences describing publicly traded companies and has become the most widely used benchmark for financial sentiment classification. Each sentence was independently labeled by sixteen annotators with financial expertise as positive, neutral, or negative.

Following prior studies, we use the *Sentences_AllAgree* subset, which contains 2,264 sentences for which all annotators assigned the same sentiment label. Restricting the analysis to unanimous observations minimizes label ambiguity and provides a clean benchmark for evaluating model performance. The sample consists of approximately 61% neutral, 25% positive, and 13% negative sentences. Because the classes are unbalanced, we report both classification accuracy and macro F1 scores throughout the analysis.

To evaluate the economic relevance of financial sentiment measures, we construct a dataset of 221 quarterly earnings announcements obtained from SEC Form 8-K filings between 2022 and mid-2026. The sample comprises 15 large U.S. firms across seven sectors: information technology (Apple, Microsoft, Nvidia, and Intel), communication services (Alphabet and Meta), consumer discretionary (Amazon and Tesla), financials (JPMorgan Chase and Bank of America), health care (Johnson & Johnson), energy (ExxonMobil), and consumer staples (Walmart, Procter & Gamble, and Coca-Cola).

**Table 1. Descriptive Statistics of Sentiment Measures and Economic Variables**

| Variable | N | Mean | Std. Dev. | Min | Max |
|---|---|---|---|---|---|
| **Sentiment measures** | | | | | |
| VADER | 221 | 0.107 | 0.044 | −0.018 | 0.220 |
| TextBlob | 221 | 0.085 | 0.081 | −0.093 | 0.360 |
| Loughran–McDonald | 221 | 0.752 | 0.209 | 0.000 | 0.955 |
| FinBERT | 221 | 0.653 | 0.527 | −0.952 | 0.942 |
| FinBERT-Tone | 221 | 0.813 | 0.426 | −1.000 | 1.000 |
| DistilRoBERTa-Financial | 221 | 0.737 | 0.628 | −0.998 | 1.000 |
| FinancialBERT | 221 | 0.824 | 0.458 | −0.997 | 1.000 |
| RoBERTa-Large-Financial | 221 | 0.595 | 0.659 | −0.908 | 0.915 |
| LLaMA 3.2 | 221 | 0.678 | 0.424 | −0.870 | 1.000 |
| Gemma 3 | 221 | 0.790 | 0.255 | −0.805 | 0.935 |
| Qwen 2.5 | 221 | 0.704 | 0.399 | −0.805 | 1.000 |
| Mistral | 221 | 0.642 | 0.548 | −0.935 | 0.974 |
| **Economic variables** | | | | | |
| EPS surprise (%) | 221 | 33.914 | 347.351 | −1,533.520 | 3,121.290 |

| Next-day return | 221 | −0.001 | 0.062 | −0.261 | 0.244 |
|---|---|---|---|---|---|

*Notes*: This table reports descriptive statistics for the sentiment scores generated by the 12 sentiment models and the two economic outcome variables. Sentiment scores are computed from 221 quarterly earnings announcements. Dictionary-based methods produce scores on different scales than transformer-based and large language models; therefore, mean values are not directly comparable across models. EPS surprise is measured as the percentage difference between reported and consensus analyst earnings forecasts. Next-day return is the stock's raw return on the first trading day following the earnings announcement.

The firms were purposively selected because their stocks are highly liquid, they receive extensive analyst and media coverage, and their earnings announcements are closely followed by market participants. These characteristics provide a useful setting for examining whether sentiment measures capture economically meaningful information in widely scrutinized corporate disclosures. However, the sample is illustrative rather than representative of the broader U.S. economy. Accordingly, the findings may not generalize to smaller firms, less-liquid securities, or companies with limited analyst coverage.

For each earnings announcement, we extract the associated earnings press release from Exhibit 99.1 of the corresponding SEC filing. We then match each announcement with two measures of firm performance: the reported earnings-per-share (EPS) surprise and the firm's next-day stock return. The EPS surprise is calculated as the percentage difference between the reported EPS and the consensus EPS forecast obtained from Yahoo Finance. Next-day stock return is measured using the percentage change in the adjusted closing stock price from the announcement day to the following trading day. This matching procedure allows us to evaluate whether model-generated sentiment is associated with underlying firm fundamentals, investor reactions, or both. If a sentiment measure captures economically relevant information, it should be associated with earnings surprises, next-day stock returns, or both. Table 1 provides descriptive statistics for the sentiment measures and economic variables.

**4. Methodology**

We evaluate three broad classes of sentiment measures that represent successive generations of natural language processing methods. The first consists of dictionary-based approaches, including VADER, TextBlob, and the Loughran–McDonald financial dictionary. These methods assign sentiment based on predefined word lists and lexical rules. The second class includes finance-specific transformer models trained or fine-tuned for financial text. We evaluate FinBERT, FinBERT-Tone, FinancialBERT, DistilRoBERTa-Financial, and RoBERTa-Large-Financial.

Unlike dictionary methods, transformer models account for word order and contextual information, allowing them to interpret complex financial language. The third class consists of modern open-source large language models (LLMs), including Llama 3.2, Gemma 3, Qwen 2.5, and Mistral. These models are evaluated without task-specific fine-tuning using a common prompt that instructs each model to classify financial text as positive, neutral, or negative.

Our objective is not simply to compare classification accuracy but to determine whether different sentiment measures capture economically meaningful information. Accordingly, we evaluate each model along two complementary dimensions. Figure 1 summarizes our conceptual framework underlying our evaluation strategy.

The first dimension assesses linguistic validity using the Financial PhraseBank. Specifically, we compare model predictions with expert human annotations and report classification accuracy, macro F1 scores, and class-specific F1 scores. Because all models are evaluated on the same set of observations, differences in classification accuracy are assessed using McNemar's test (McNemar, 1947).

The second dimension evaluates economic validity using our earnings announcement dataset. For each earnings press release, we examine the association between model-generated sentiment, earnings surprises, and next-day stock returns. We compute Spearman rank correlations because neither sentiment scores nor earnings surprises are normally distributed. Statistical significance is assessed using bootstrap confidence intervals and conventional hypothesis tests.

**Figure 1. Conceptual Framework**

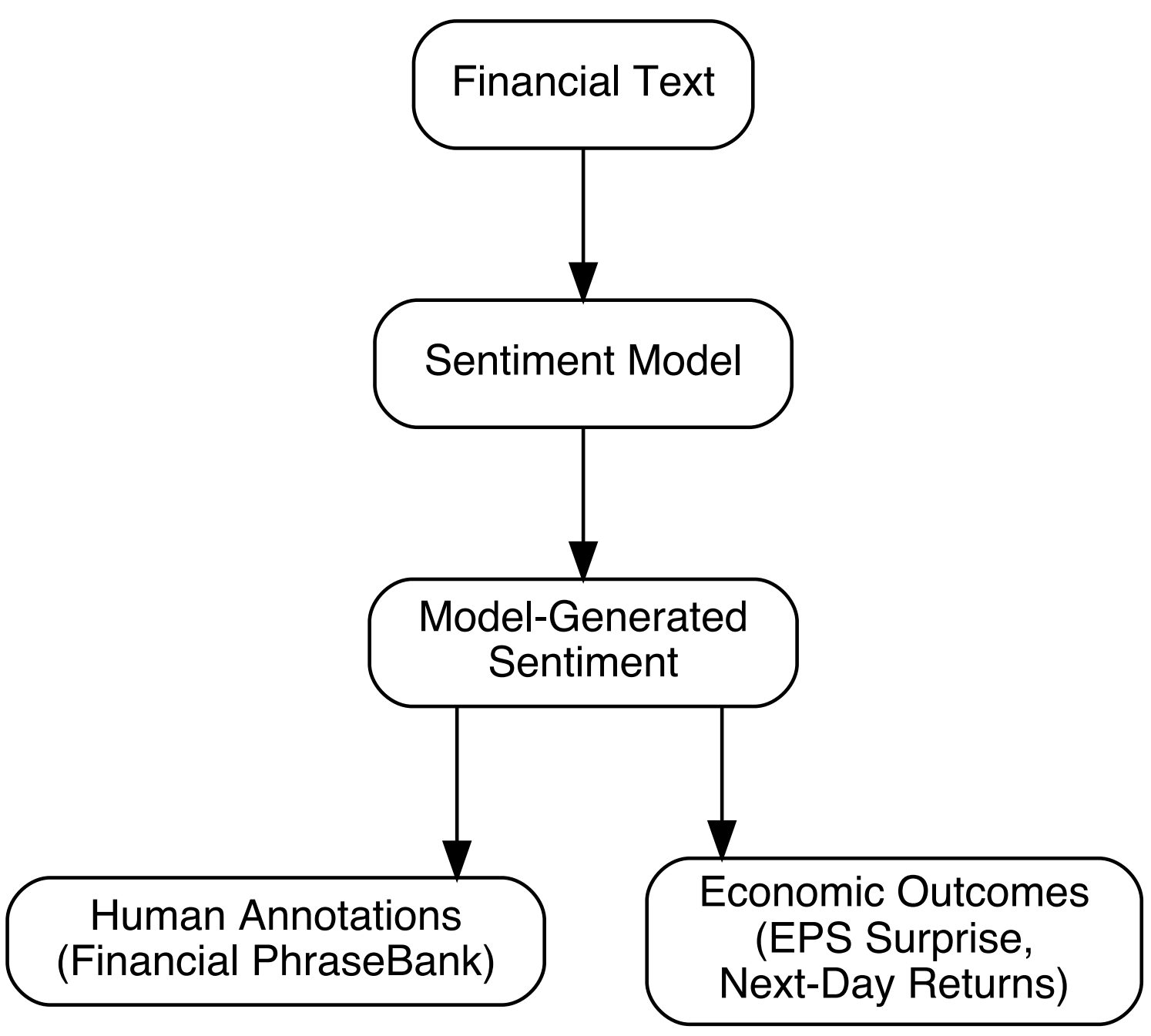

## 5. Empirical Results

Table 2 summarizes the results of sentiment models to reproduce expert human judgments using the Financial PhraseBank. Consistent with prior studies, finance-specific transformer models substantially outperform traditional dictionary-based approaches. Dictionary methods achieve classification accuracies ranging from 54% to 65%, while all finance-specific transformer models achieve accuracies above 90%. Among the transformer models, RoBERTa-Large-Financial and DistilRoBERTa-Financial achieve the highest classification accuracies. However, these results should be interpreted with caution because both models were fine-tuned using the Financial PhraseBank. Consequently, their near-perfect performance partly reflects overlap between the evaluation dataset and their training data.

**Table 2. Classification Performance on the Financial PhraseBank**

| Model | Category | Accuracy | Macro F1 | Pos F1 | Neg F1 | Neu F1 |
|---|---|---|---|---|---|---|
| TextBlob | Lexicon | 54.3 | 46.7 | 35.8 | 39.1 | 65.2 |
| VADER | Lexicon | 57.1 | 48.7 | 51.0 | 27.8 | 67.3 |
| Loughran-McDonald | Lexicon | 64.3 | 51.2 | 45.9 | 32.1 | 75.8 |
| Gemma 3 (1B) | LLM | 73.0 | 73.1 | 67.5 | 76.0 | 75.7 |
| LLaMA 3.2 (3B) | LLM | 82.3 | 81.8 | 78.3 | 82.8 | 84.3 |
| Mistral (7B) | LLM | 88.7 | 88.6 | 82.7 | 92.4 | 90.7 |
| FinBERT-Tone | Transformer | 91.7 | 89.7 | 83.8 | 90.6 | 94.7 |
| Qwen 2.5 (3B)[a] | LLM | 92.1 | 90.7 | 86.2 | 91.5 | 94.4 |
| FinBERT | Transformer | 97.2 | 96.3 | 96.2 | 94.3 | 98.2 |
| FinancialBERT | Transformer | 98.9 | 98.6 | 98.0 | 98.5 | 99.4 |
| DistilRoBERTa-Financial | Transformer | 99.7 | 99.6 | 99.7 | 99.2 | **99.9** |
| RoBERTa-Large-Financial[a] | Transformer | **99.9** | **99.9** | **99.8** | **99.8** | **99.9** |

*Notes*: Results are based on the Financial PhraseBank (sentences_allagree, n = 2,264). All values are percentages. Bold denotes the best-performing model in each column. A superscript *a* indicates that the model's classification accuracy is not significantly different from the model immediately above based on McNemar's test ($p > 0.05$).

The more informative comparison is between general-purpose LLMs and finance-specific models that were not trained on the Financial PhraseBank. Qwen 2.5 achieves 92.1% classification accuracy, which is statistically indistinguishable from the 91.7% accuracy of FinBERT-Tone. This result suggests that recent advances in general-purpose LLMs have substantially reduced the performance gap between specialized financial language models and models trained on general text.

Figure 2 reports model error rates across different sentence characteristics, including negation, uncertainty, financial jargon, and sentence length. Dictionary-based methods exhibit substantially higher error rates across all sentence categories. In contrast, finance-specific transformer models maintain consistently low error rates, while general-purpose LLMs substantially outperform dictionary-based methods but generally exhibit higher error rates than the finance-specific transformers. As noted earlier, however, several finance-specific transformer models were fine-tuned using the Financial PhraseBank. Consequently, their superior performance should be interpreted with caution, as it may partly reflect prior exposure to the benchmark dataset during fine-tuning.

Overall, the results indicate that modern LLMs provide a practical alternative to finance-specific sentiment models for financial text classification. Although transformer models continue to outperform dictionary-based methods, the differences between the strongest finance-specific models and general-purpose LLMs are relatively modest.

**Figure 2. Error Rates by Sentence Type**

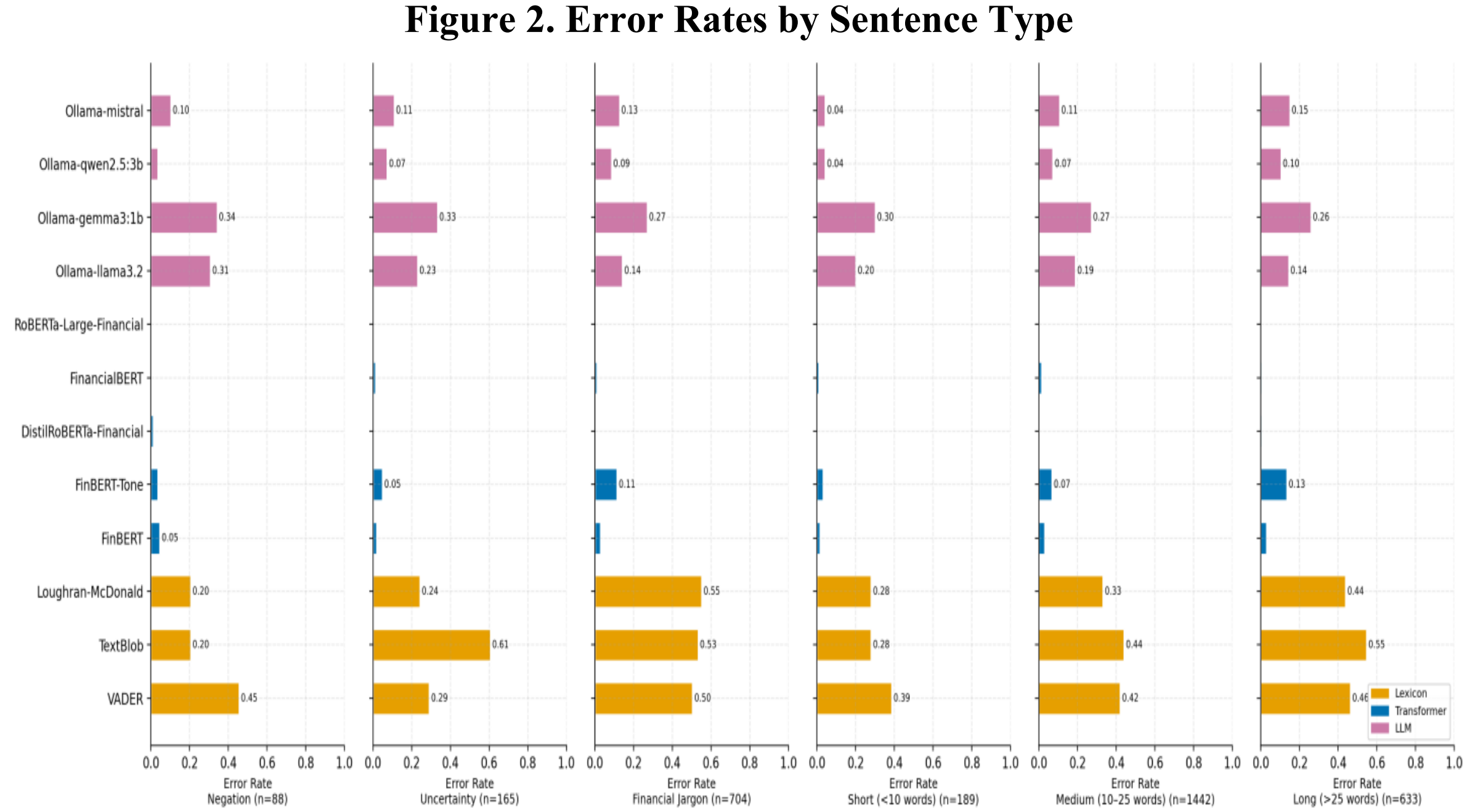

*Notes*: Each sentence in the Financial PhraseBank is assigned to six categories using rule-based heuristics. The three linguistic categories are negation, uncertainty, and financial jargon. These categories are identified using predefined keyword lists consisting of 27 negation terms, 22 uncertainty terms, and 27 finance-specific terms, respectively. A sentence is assigned to a linguistic category if it contains at least one keyword from the corresponding list and may therefore receive multiple linguistic tags. In addition, sentence length is measured by the number of tokens and classified into one of three mutually exclusive categories: short, medium, or long. The keyword lists are heuristic rather than manually validated annotations and are included in the replication files.

Although overall classification accuracy provides a useful benchmark, it does not necessarily indicate whether a sentiment measure captures economically meaningful information. We therefore examine the relationship between model-generated sentiment and firm fundamentals. Figure 3 reports the results. Across all twelve sentiment models, none exhibits a statistically significant association with next-day stock returns. Estimated correlations are generally positive but economically small and statistically indistinguishable from zero. These findings suggest that improvements in benchmark classification accuracy do not translate into stronger predictive relationships with short-run stock returns.

The results differ when earnings surprises are used as the validation target. Several finance-specific transformer models exhibit statistically significant positive correlations between sentiment and earnings surprises, with FinancialBERT producing the strongest association. Among the general-purpose LLMs, Mistral and Llama 3.2 also exhibit positive correlations with earnings surprises, although the estimates are less precise.

To better understand these results, we conduct an exploratory analysis by dividing the 221 earnings announcements into quartiles based on the magnitude of the EPS surprise and computing the directional accuracy and Spearman's ρ within each quartile. Figure 4 shows that model performance varies systematically with the magnitude of the earnings surprise. Both directional accuracy and Spearman's ρ are highest in the lowest and highest EPS surprise quartiles, corresponding to announcements with the largest earnings misses and the largest earnings beats, respectively. In contrast, performance is substantially weaker in the middle quartiles, where earnings surprises are relatively small. Although there is some variation across individual models, the overall pattern is remarkably consistent across sentiment measures. These findings suggest that financial sentiment models are most effective at detecting sentiment when earnings announcements convey clearly positive or clearly negative information. When reported earnings are close to market expectations, the language of the accompanying disclosures is likely to be more nuanced, making it more difficult for both traditional sentiment models and LLMs to distinguish positive from negative sentiment accurately.

**Figure 3. Spearman Rank Correlations Between Model-generated Sentiment, Earnings Surprises, and Next-Day Stock Returns**

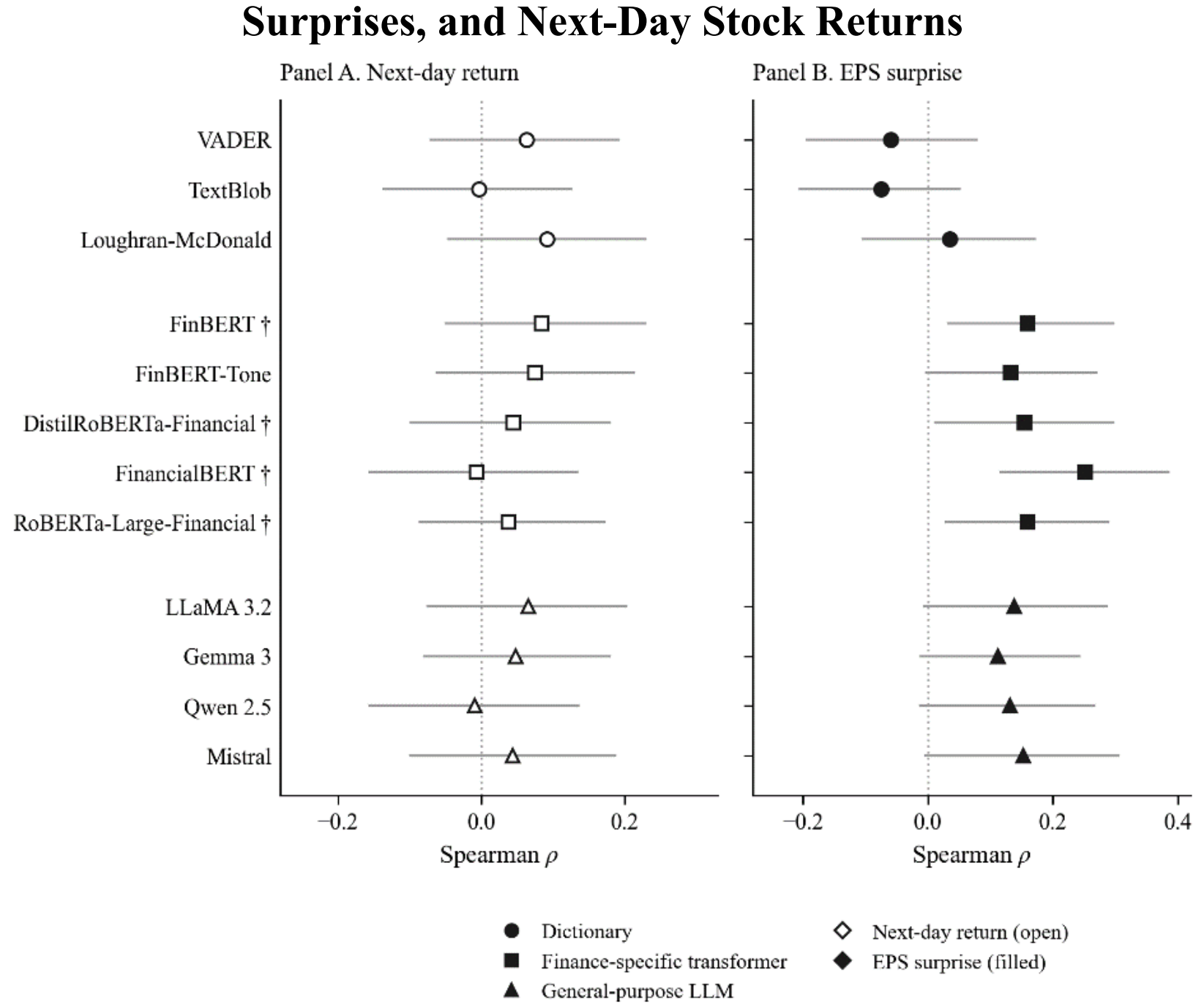

*Notes*: The sample consists of 221 earnings announcement events. Correlations are reported with 95% bootstrap confidence intervals based on 2,000 event-level bootstrap resamples.

**Figure 4. Directional Accuracy (Left) and Spearman ρ (Right) by EPS Surprise Quartile**

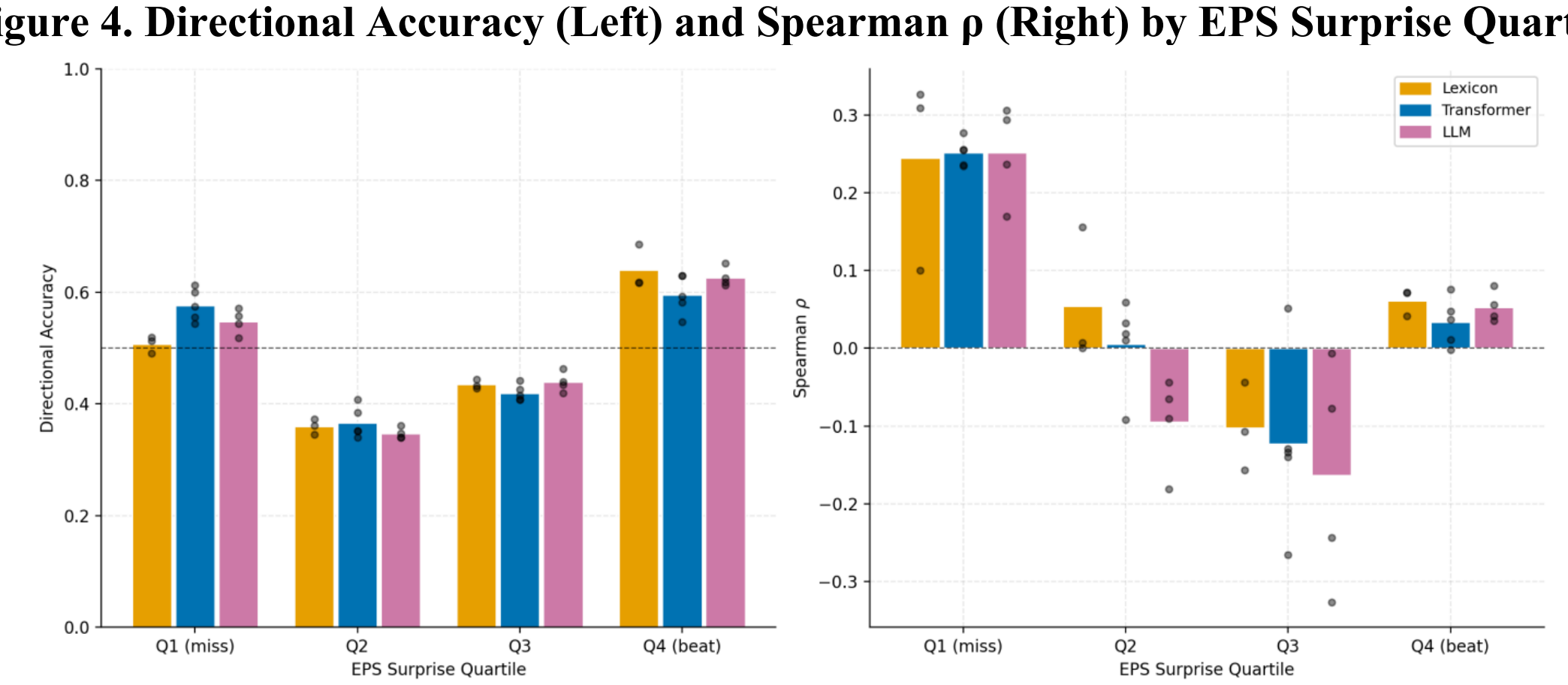

*Notes*: Bars represent quartile means, and dots represent individual sentiment models.

Taken together, these findings demonstrate that linguistic performance and economic usefulness are distinct concepts. Models that more closely reproduce expert sentiment labels do not necessarily provide more informative measures of firms' underlying economic performance or short-run market reactions. Instead, sentiment measures appear to be most informative when firms report substantial positive or negative earnings surprises, suggesting that the economic value of sentiment depends not only on model quality but also on the information content of the underlying disclosure.

## 6. Additional Analysis

To examine whether access to a proprietary frontier model materially changes the conclusions, we conduct an additional sensitivity analysis using Claude Opus 4.5. Claude represents one of the most advanced commercially available large language models and serves as a useful benchmark against the open-source LLMs evaluated in this study. We apply the same prompting strategy and evaluation procedures used for the other models, allowing a direct comparison of linguistic accuracy and economic validity.

Table 3 shows that Claude achieves an accuracy of 93.6% and a macro F1 score of 93.3% on the Financial PhraseBank. Although Claude performs strongly across all sentiment classes, its classification accuracy is not statistically different from that of Qwen 2.5, an open-source LLM, according to the McNemar test ($p > 0.05$). This finding suggests that recent open-source LLMs have largely closed the performance gap with leading proprietary models on benchmark sentiment classification.

**Table 3. Comparison of Open-Source and Closed-Source Large Language Models on the Financial PhraseBank**

| Model | Category | Accuracy | Macro F1 | Pos F1 | Neg F1 | Neu F1 |
|---|---|---|---|---|---|---|
| TextBlob | Dictionary | 54.3 | 46.7 | 35.8 | 39.1 | 65.2 |
| VADER | Dictionary | 57.1 | 48.7 | 51.0 | 27.8 | 67.3 |
| Loughran-McDonald | Dictionary | 64.3 | 51.2 | 45.9 | 32.1 | 75.8 |
| Gemma 3 (1B) | General-purpose | 73.0 | 73.1 | 67.5 | 76.0 | 75.7 |
| LLaMA 3.2 (3B) | General-purpose | 82.3 | 81.8 | 78.3 | 82.8 | 84.3 |
| Mistral (7B) | General-purpose | 88.7 | 88.6 | 82.7 | 92.4 | 90.7 |
| FinBERT-Tone | Finance-specific | 91.7 | 89.7 | 83.8 | 90.6 | 94.7 |
| Qwen 2.5 (3B)[a] | General-purpose | 92.1 | 90.7 | 86.2 | 91.5 | 94.4 |
| Claude Opus 4.5[a] | Commercial | 93.6 | 93.3 | 90.9 | 94.3 | 94.6 |
| FinBERT | Finance-specific | 97.2 | 96.3 | 96.2 | 94.3 | 98.2 |
| FinancialBERT | Finance-specific | 98.9 | 98.6 | 98.0 | 98.5 | 99.4 |
| DistilRoBERTa-Financial | Finance-specific | 99.7 | 99.6 | 99.7 | 99.2 | **99.9** |
| RoBERTa-Large-Financial[a] | Finance-specific | **99.9** | **99.9** | **99.8** | **99.8** | **99.9** |

Notes: Claude Opus 4.5 is included as a representative proprietary frontier large language model. Performance is evaluated on the Financial PhraseBank (sentences_allagree, n = 2,264). All values are percentages. Bold indicates the best value in each column. A superscript a denotes that the model is statistically tied with the row above according to McNemar's test ($p > 0.05$). Pos, Neg, and Neu F1 are class-specific F1 scores.

Figure 5 indicates that Claude's sentiment exhibits a positive association with earnings surprises, while, consistent with the other models, it shows no statistically significant association with next-day stock returns. Overall, Claude produces results that are qualitatively similar to those of the open-source LLMs and does not alter the main conclusions of the paper.

**Figure 5. Economic Validity of Sentiment Measures Including Claude Opus 4.5**

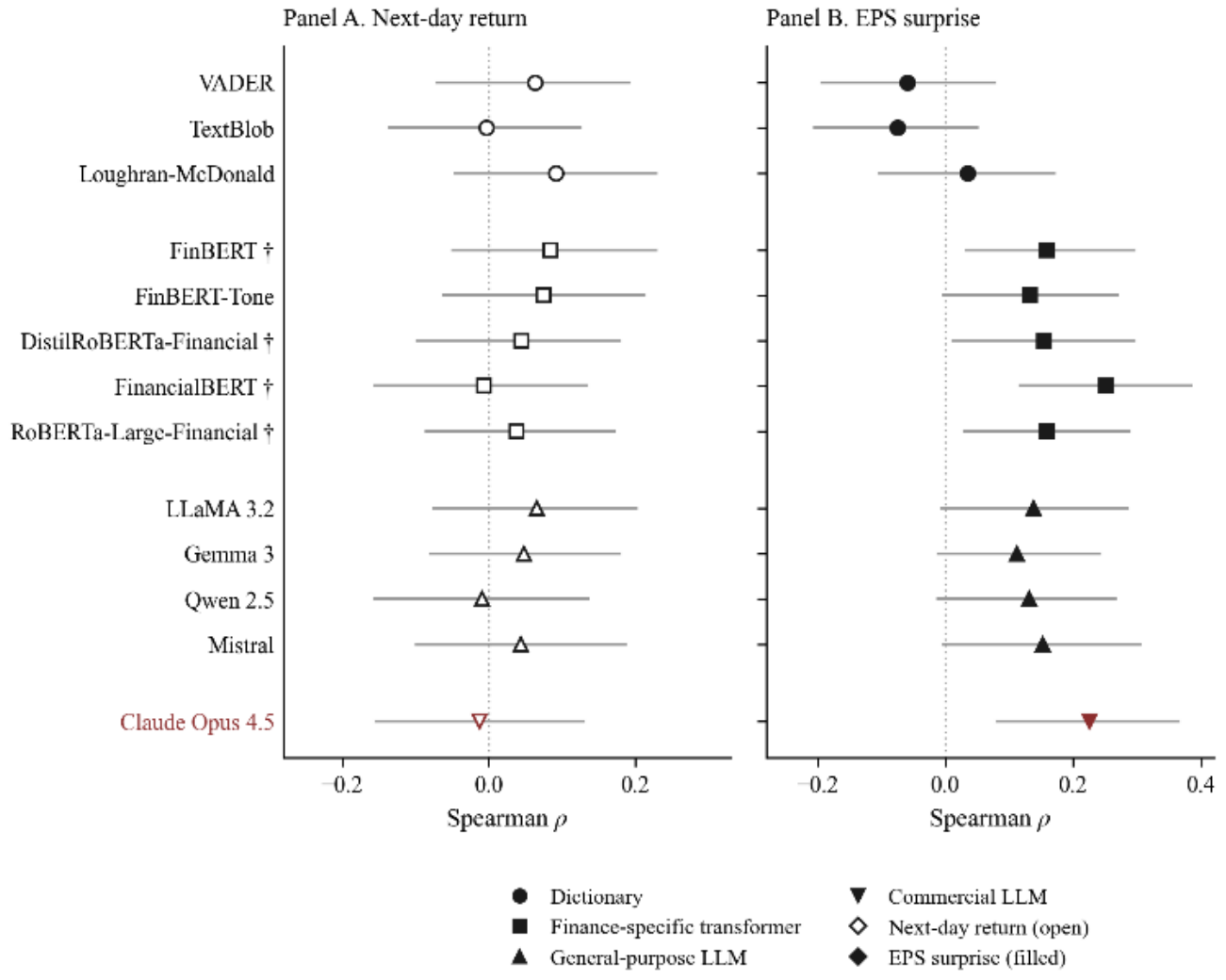

Notes: The figure reports Spearman rank correlations (ρ) between model-generated sentiment and (Panel A) next-day stock returns and (Panel B) earnings surprises for 221 earnings announcements. Points represent the estimated correlation, and horizontal bars denote 95% confidence intervals.

The findings indicate that proprietary frontier models provide little incremental benefit over modern open-source LLMs for financial sentiment analysis. Claude exhibits a positive association with earnings surprises similar to the open-source models and, like all other models considered in this study, is not significantly associated with next-day stock returns. These results reinforce the paper's central conclusion that open-source LLMs provide a cost-effective and transparent alternative to commercial systems without sacrificing either linguistic or economic performance.

## 7. Discussion

Our findings demonstrate that improvements in benchmark classification accuracy do not necessarily translate into more economically informative measures of financial sentiment. While several finance-specific transformer models achieve near-perfect agreement with expert annotations, none exhibits a statistically significant association with next-day stock returns. These results highlight an important distinction between measuring linguistic performance and measuring economic relevance.

One explanation is that textual sentiment represents only one component of the information contained in corporate disclosures. Earnings announcements include both qualitative language and quantitative information, such as earnings, revenues, and forward-looking guidance. Investors incorporate all available information when valuing firms, making it unlikely that textual sentiment alone can explain a substantial portion of short-run stock price movements. Consequently, even highly accurate sentiment models should not be expected to generate strong return predictability.

At the same time, our additional analysis indicates that financial sentiment is more informative when firms report extreme earnings outcomes. Sentiment measures exhibit substantially higher directional accuracy and stronger rank correlations for announcements associated with the largest earnings beats and misses than for announcements with relatively small earnings surprises. This pattern suggests that the qualitative tone of earnings disclosures more clearly reflects underlying firm performance when economic news is unambiguously positive or negative. In contrast, announcements with modest earnings surprises often contain mixed signals, making sentiment more difficult to distinguish and less informative. These findings imply that financial sentiment may be most valuable in settings where firms experience large fundamental shocks rather than routine earnings announcements.

Our findings are also consistent with the presence of measurement error. Financial sentiment is a latent construct that cannot be directly observed. Human annotations provide one measure of sentiment, but they do not necessarily represent the information that investors consider most relevant. Likewise, sentiment models estimate tone from text, introducing additional measurement error. Because earnings surprises themselves exhibit only a modest association with next-day stock returns, any attenuation in measuring sentiment further weakens the relationship between sentiment and market outcomes.

The results have practical implications for empirical finance research. Finance-specific transformer models remain attractive when the primary objective is to reproduce expert sentiment labels. However, modern open-source LLMs achieve comparable performance without task-specific fine-tuning while offering substantially greater flexibility for other language tasks, including summarization, information extraction, and question answering. Researchers already using LLMs for multiple aspects of text analysis may therefore find that a single general-purpose model is sufficient for measuring financial sentiment as well.

More broadly, our results suggest that benchmark datasets should not be the sole basis for selecting sentiment models. Benchmark evaluations measure agreement with expert judgments, whereas empirical finance applications seek measures that capture economically meaningful information. Researchers should therefore validate sentiment measures against outcomes relevant to their research questions rather than relying exclusively on benchmark classification performance.

## 8. Conclusion

Financial sentiment measures have become an important tool in empirical finance, yet relatively little evidence exists on whether recent advances in large language models improve their usefulness for economic research. This paper evaluates twelve sentiment models, including dictionary-based methods, finance-specific transformer models, and general-purpose LLMs, using both expert human annotations and real-world earnings announcements.

Our findings lead to three main conclusions. First, modern general-purpose LLMs achieve classification performance comparable to finance-specific language models despite requiring no task-specific fine-tuning. Second, improvements in benchmark classification accuracy do not translate into stronger relationships with economically meaningful outcomes. Although several sentiment measures are significantly associated with earnings surprises, none exhibits a statistically significant relationship with next-day stock returns. Third, we also find that sentiment measures are most informative for announcements with the largest earnings beats and misses, suggesting that textual sentiment more clearly reflects firm fundamentals when economic news is unambiguously positive or negative.

The broader implication is that researchers should evaluate sentiment models according to the objectives of their studies. Models designed to reproduce expert sentiment labels are appropriate when linguistic accuracy is the primary concern. However, applications in empirical finance require measures that capture economically meaningful information. Our findings further suggest that modern open-source LLMs provide a practical alternative to proprietary models, offering comparable performance while providing greater transparency, flexibility, and lower implementation costs. Future evaluations of financial sentiment models should therefore combine benchmark datasets with economic validation rather than relying exclusively on classification performance.

This study also suggests several directions for future research. First, future work should examine whether these findings extend to other forms of financial text, including analyst reports, conference call transcripts, and regulatory filings. Second, alternative measures of economic relevance, such as abnormal returns, analyst forecast revisions, or changes in market expectations, may provide additional insights into the value of financial sentiment measures. Finally, as large language models continue to evolve, future research should examine whether improvements in general-purpose language models translate into meaningful gains for empirical finance applications.

Overall, our results indicate that advances in language models have substantially improved the measurement of financial sentiment. However, benchmark classification accuracy alone should not be viewed as evidence of greater economic usefulness. The value of a financial sentiment measure ultimately depends on its ability to capture information relevant to the economic question being studied.

**Disclosures**

- Disclosure of interest: The authors declare that they have no competing interests.
- Funding: No funding was received.